\documentclass[journal]{IEEEtran}
\usepackage{amsmath,amsfonts}
\usepackage{array}
\usepackage[caption=false,font=normalsize,labelfont=sf,textfont=sf]{subfig}
\usepackage{textcomp}
\usepackage{stfloats}
\usepackage{url}
\usepackage{verbatim}
\usepackage{graphicx}
\usepackage{gensymb}
\usepackage{tikz}
\usepackage[table]{xcolor}
\usetikzlibrary{arrows.meta, positioning}
\usepackage{cite}
\usepackage{float}
\usepackage{comment}
\usepackage{booktabs}
\usepackage{multirow}
\usepackage[ruled,vlined]{algorithm2e}
\begin{document}

\title{Cooperative Multi-UAV Navigation in Complex Environments via Systematic Multi-Agent Deep Reinforcement Learning}

\author{\IEEEauthorblockN{Yu Su and Nabil Aouf}\\
\IEEEauthorblockA{School of Science and Technology\\
City St George's University of London\\
London, United Kingdom\\
Email: Yu.Su@city.ac.uk, nabil.aouf@city.ac.uk}

}

\maketitle

\begin{abstract}

Cooperative navigation of multi-agent UAVs in complex environments faces key challenges such as local optima traps, sparse rewards, learning imbalance among agents, and insufficient cross-scenario generalisation capabilities. This paper proposes a multi-agent deep reinforcement learning framework that systematically addresses these issues through coordinated exploration, demonstration exploitation, safe curriculum scheduling, and structure-aware generalisation mechanisms.

Firstly, the proposed perception mechanism combines memory of previously visited states, estimates of directional novelty, and backpropagation of penalties, enabling the agent to proactively detect and escape from local optima. Secondly, the hierarchical collaborative demonstration buffer and the tiered behaviour cloning scheme manage trajectories in layers based on the degree of team collaboration, and apply differential supervision directly to the actor network, thereby enhancing the efficiency of demonstration utilisation under sparse collaborative signals. Furthermore, the safety-aware dual-condition curriculum scheduling mechanism, whilst advancing the training phase, actively reviews mastered scenarios through back-testing and experience pre-filling, effectively suppressing catastrophic forgetting and achieving concurrent assurance of both task performance and flight safety. Regarding generalisation, local geometric features deterministically computed from sensor readings are abstracted into a domain parameter, through which a structure-aware gating network and mixture-of-experts mechanism condition the policy on local structural patterns rather than scenario-specific coordinate memory, enabling cross-scenario transfer without exposure to the target environment.

The framework is further validated under a mixed static–dynamic obstacle setting, demonstrating robust adaptability to dynamic environmental disturbances. Simulation results confirm strong performance across collaboration success rate, navigation robustness, zero-shot cross-scenario generalisation, and dynamic environment adaptability.

\end{abstract}


\section{Introduction}

\subsection{Research Background}

In applications such as disaster response, industrial inspection, and warehouse automation, multiple UAVs are required to perform cooperative navigation in complex and constrained environments typically characterised by intricate spatial structures, severe occlusions, and the absence of global positioning information. Under these conditions, agents are prone to becoming trapped by local structures, whilst dynamic interactions among multiple agents introduce non-stationarity and potential learning imbalance. Enabling efficient exploration under partial observability, avoiding local optima, and improving coordinated decision-making therefore remain key challenges in multi-UAV autonomous navigation.

Intrinsic motivation-based exploration strategies improve learning efficiency by encouraging agents to visit unseen states. Curiosity-driven methods use state-transition prediction errors as intrinsic rewards~\cite{pathak2017curiosity}, novelty-based approaches quantify state rarity via random network distillation~\cite{burda2018exploration}, and memory-based methods incorporate visitation history and multi-scale novelty signals to reduce redundant behaviour~\cite{badia2020never}. Whilst these methods promote broader state coverage, they lack explicit task-relevant guidance and offer no targeted mechanism for escaping local optima in structurally constrained environments.

To improve spatial awareness and long-term planning, map-based and memory-driven navigation methods have been widely adopted. Occupancy grid mapping provides probabilistic environmental representations for structured spatial reasoning~\cite{thrun2002probabilistic}, neural map representations capture spatial dependencies in partially observable settings~\cite{parisotto2017neural}, and coverage-driven strategies leverage visitation history to reduce redundant exploration~\cite{chaplot2020object}. However, these methods focus on environmental representation rather than active intervention, and agents may still converge to suboptimal paths in complex maze structures.

Multi-agent coordination methods address interaction and cooperation in shared environments. Centralised training with decentralised execution (CTDE)-based approaches improve training stability by allowing critics to access global state whilst maintaining decentralised execution~\cite{lowe2017multi}, communication-based methods enable information exchange for coordinated decision-making~\cite{foerster2016learning}, and value factorisation approaches decompose global value functions to improve joint policy optimisation~\cite{rashid2020monotonic}. Nevertheless, these methods remain susceptible to systemic stagnation under strong inter-agent coupling and provide no targeted response to learning imbalance across agents.

In policy generalisation and cross-scenario transfer, domain randomisation improves robustness through diverse environment perturbations during training~\cite{tobin2017domain}, whilst meta-learning enables rapid adaptation from limited interaction samples~\cite{finn2017model}. Context-conditioned policy methods abstract structural or dynamic features into domain parameters to support deployment in new scenarios without retraining~\cite{liang2026causality, beukman2023dynamics}, and mixture-of-experts models blend sub-policy outputs through a gating mechanism to improve multi-task adaptability~\cite{obando2024mixtures}. However, these methods primarily target generalisation over physical dynamics or task objectives, with little attention given to environments characterised by geometric repeatability.

These limitations collectively motivate the development of a framework that coordinates multi-agent training and structure-aware generalisation under sparse reward conditions, enabling reliable collaborative drone navigation in complex structural environments characterised by dynamic obstacles and local optima traps.

\subsection{Previous Work}

\subsubsection{Intrinsic Motivation-based Exploration Methods}
Curiosity-driven methods use state-transition prediction errors as intrinsic exploration signals~\cite{pathak2017curiosity}, random network distillation measures state novelty by comparing the outputs of a fixed randomly initialised network and a predictor network~\cite{burda2018exploration}, and count-based or pseudo-count methods assign higher intrinsic rewards to infrequently visited states to promote broader state-space coverage~\cite{bellemare2016unifying}. Whilst these approaches have demonstrated effectiveness in sparse-reward settings, the decoupling of intrinsic rewards from task objectives may direct agents towards task-irrelevant regions; in structurally constrained environments with multiple suboptimal paths, agents may remain trapped in local stagnation despite strong exploratory capability.

\subsubsection{Map-based and Memory-driven Navigation Methods}
Map-based methods typically combine classical planning algorithms such as A*~\cite{hart1968formal} or Dijkstra~\cite{dijkstra2022note} to achieve stable path planning and obstacle avoidance in reconstructable environments, and visibility-graph extensions have been proposed for 3D constrained UAV path planning~\cite{blasi2022uav}; however, these approaches require complete environmental prior knowledge and cannot support real-time multi-agent coordination. Memory-driven methods encode historical observations via recurrent neural networks or external memory modules; the Neural Map approach~\cite{parisotto2017neural} builds spatial cognition through learnable memory structures, whilst attention-augmented memory models further improve the extraction of salient historical information~\cite{parisotto2017neural}. These methods rely on relatively accurate environmental modelling and incur substantial computational overhead for map construction and maintenance; more fundamentally, their focus on environmental representation rather than active intervention leaves agents without an effective mechanism for escaping suboptimal paths in complex maze structures.

\subsubsection{Multi-agent Coordination and Deadlock Avoidance}
Value decomposition methods such as VDN~\cite{sunehag2017value} and QMIX~\cite{rashid2020monotonic} factorise the global value function into local value functions to model joint behaviour, whilst policy-gradient methods introduce a centralised critic to improve training stability~\cite{lowe2017multi}. In terms of specific coordination mechanisms, existing work commonly employs cooperative reward design, conflict penalties, or priority-based action scheduling to reduce collision and deadlock rates. Nevertheless, under strong inter-agent coupling, these methods remain prone to stable yet suboptimal behavioural patterns such as persistent loops or mutual path blocking; moreover, when individual agents lag significantly in learning progress, existing approaches lack dedicated mechanisms for addressing inter-agent learning imbalance, limiting overall cooperative efficiency~\cite{azzam2023learning,wan2025imitation}.

\subsubsection{Policy Generalisation and Cross-scenario Transfer}
Domain randomisation~\cite{tobin2017domain} and meta-learning~\cite{finn2017model} improve policy generalisation through training diversity and rapid task adaptation, respectively. Context-conditioned policy methods abstract structural or dynamic environment features into domain parameters to support cross-scenario deployment without retraining~\cite{liang2026causality,beukman2023dynamics}, and mixture-of-experts models blend multiple sub-policy outputs via a gating mechanism to improve adaptability across multi-task settings~\cite{obando2024mixtures}. However, these methods target generalisation over variations in physical dynamics or task objectives; the structural regularity arising from recurring local geometric patterns in maze-like environments has received little targeted treatment, and local structure perception has not been integrated into multi-agent cooperative training, limiting effective cross-scenario transfer in such settings~\cite{liu2026self,tong2025rapid}.

\subsection{Present Work}

Achieving reliable cooperative navigation in structurally constrained environments remains challenging due to the simultaneous presence of local optima, sparse cooperative success signals, and learning imbalance across agents. Whilst prior work has addressed individual aspects of this problem, intrinsic motivation methods decouple exploration signals from task objectives, map-based methods offer no active intervention for trapped agents, and multi-agent coordination methods remain susceptible to systemic stagnation under strong inter-agent coupling. No existing approach provides a unified framework that jointly resolves these limitations, and their compounding interaction in complex maze environments constitutes the central motivation of the present work.

To address the shared deficiency of intrinsic motivation and map-based methods in handling local optima, this work proposes an online local-optima diagnosis and graded behavioural intervention mechanism that is explicitly conditioned on the task objective. Rather than superimposing an auxiliary exploration signal onto the policy gradient, the proposed mechanism operates directly at the action-execution layer, applying independent real-time intervention to each agent without introducing any additional trainable parameters. By integrating visited-state memory, directional novelty estimation, and movement-history analysis, the mechanism identifies local stagnation with high precision and, upon detection, overrides the policy output to actively steer the agent towards unvisited regions. To further address the reward vacuum that conventional distance-based shaping creates when an agent is forced to detour through a corridor, a LiDAR-derived passage-constraint indicator is introduced to dynamically adjust the reward structure during occlusion phases, ensuring that meaningful learning signal is preserved even when the goal is obstructed. The mechanism is parameter-free and operates independently for each agent, rendering it scalable to an arbitrary number of vehicles.

To remedy the fundamental limitation of existing LfD~\cite{argall2009survey} and GAIL~\cite{ho2016generative} approaches, which treat all demonstration data homogeneously without regard to cooperative task completion, this work constructs a hierarchical demonstration buffer in which trajectories are stratified according to team-level task achievement. Trajectories corresponding to full cooperative arrival and partial arrival are stored in separate buffers, each governed by a fine-grained quota management scheme that operates at both the scenario level and the start–goal pair level, ensuring that high-quality cooperative trajectories across varying difficulty conditions are retained in a balanced manner. Behavioural cloning supervision is applied directly to the actor network with differentiated weighting, eliminating the cumulative approximation error introduced by the two-stage policy-distillation paradigm that relies on an auxiliary network as an intermediary. This design encodes cooperative completion degree as a first-class attribute of the demonstration data, and addresses learning imbalance among agents at both the data-management and policy-optimisation levels.

To overcome the susceptibility of existing multi-agent coordination methods to systemic stagnation and catastrophic forgetting across training scenarios, this work introduces a safety-aware dual-condition curriculum scheduling mechanism. Curriculum advancement requires both a cooperative success rate threshold and a collision rate ceiling to be satisfied simultaneously, preventing the policy from exploiting high-risk behaviours to achieve superficially favourable success metrics and thereby enforcing concurrent guarantees on task performance and flight safety. Upon passing each stage, the framework immediately conducts a retrospective full evaluation over all previously mastered scenarios and proactively prefills the experience replay buffer with freshly collected trajectories from those scenarios, providing a principled countermeasure against catastrophic forgetting. At the reward level, a synchronisation-aware team reward is introduced that activates cooperative bonuses only when all agents are concurrently within the goal neighbourhood, and imposes explicit penalties proportional to inter-agent arrival time disparities, encouraging the emergence of coordinated approach behaviour. This stands in contrast to existing CTDE approaches, which rely passively on a centralised critic's access to global state information to foster cooperation, and extends naturally to teams of arbitrary size through the use of intra-group arrival time variance as the synchronisation penalty signal.

The present work proposes a cooperative multi-UAV navigation framework for partially observable complex environments, built upon execution-level local-optima escape, hierarchical cooperative demonstration utilisation, direct behavioural cloning supervision, and safety-aware training scheduling. The framework systematically addresses agent stagnation in structurally constrained environments, inefficient exploitation of demonstration data under sparse cooperative success signals, and learning imbalance and catastrophic forgetting during multi-agent training, providing a principled and scalable solution to the problem of autonomous cooperative navigation in complex constrained environments.

To address the limitation of existing generalisation methods in exploiting the geometric repeatability of maze environments, this work proposes a structure-aware generalisation mechanism that abstracts the recurring local geometric structures of mazes into a domain parameter~$\omega$, computed deterministically from sensor readings at each timestep without requiring manual annotation or any prior knowledge of the current scenario. Through a structure-aware gating network and a mixture-of-experts mechanism, the policy learns behavioural strategies conditioned on local geometric features rather than scenario-specific coordinate memory, enabling direct transfer to unseen mazes composed of structural patterns encountered during training without retraining. The proposed framework is further validated in maze environments containing moving obstacles, demonstrating robust adaptability to dynamic environmental disturbances.

\subsection{Contributions}

The main contributions of this work are summarised as follows:

\begin{itemize}
\item A task-objective-aware online local-optima diagnosis and execution-level behavioural intervention mechanism is proposed, which integrates visited-state memory, directional novelty estimation, and movement-history analysis to accurately identify agent stagnation and actively override the policy output without introducing any additional trainable parameters. A LiDAR-derived passage-constraint indicator is further introduced to eliminate the reward vacuum that arises in maze corridors under conventional distance-based shaping.
\item A hierarchical demonstration buffer is proposed in which trajectories are stratified according to team-level cooperative task completion. Full cooperative arrival and partial arrival trajectories are stored in separate buffers, each governed by a fine-grained quota management scheme operating at both the scenario level and the start–goal pair level, systematically alleviating learning imbalance across agents at the data-management level.
\item A direct graded behavioural cloning scheme is proposed that supervises the actor network using cooperative and individual demonstration data with differentiated weighting, without relying on any auxiliary intermediary network. This eliminates the cumulative approximation error inherent in indirect policy-distillation paradigms and improves the efficiency of demonstration exploitation under sparse cooperative success signals.
\item A safety-aware dual-condition curriculum scheduling mechanism is proposed that requires both a cooperative success rate threshold and a collision rate ceiling to be jointly satisfied for stage advancement. Combined with retrospective full evaluation and experience prefilling, the mechanism enforces concurrent guarantees on task performance and flight safety whilst providing a principled countermeasure against catastrophic forgetting across training scenarios.
\item A structure-aware generalisation mechanism is proposed that abstracts recurring local geometric patterns into a domain parameter deterministically computed from sensor readings, through which a structure-aware gating network and mixture-of-experts mechanism condition the policy on local structural features rather than scenario-specific coordinate memory, enabling cross-scenario transfer to unseen mazes without exposure to the target environment.
\end{itemize}

\section{Methodology}
This chapter presents the complete methodological framework for the proposed cooperative multi-UAV navigation system. The problem is first formalised as a multi-agent partially observable Markov decision process, followed by a description of the overall framework architecture based on multi-agent soft actor-critic (MASAC) and the interactions among its constituent modules. The reward function design is then introduced to establish a unified learning signal foundation for all subsequent mechanisms. Building on this, the local optima diagnosis and execution-level intervention mechanism, the hierarchical cooperative demonstration buffer with graded behavioural cloning, and the safety-aware dual-condition curriculum scheduling mechanism are presented in turn, systematically addressing the core challenges of cooperative navigation from the perspectives of local optima escape, demonstration exploitation, and coordinated multi-agent training, respectively. Finally, the structure-aware generalisation mechanism and the complete training procedure are described, whereby recurring local geometric patterns are abstracted into domain parameters and combined with a mixture-of-experts architecture to enable cross-scenario transfer to unseen environments without retraining.

\subsection{Problem Formulation}

The cooperative navigation task is formalised as a decentralised partially observable Markov decision process (Dec-POMDP)~\cite{oliehoek2016concise}, defined by the tuple $\langle \mathcal{N}, \mathcal{S}, \{\mathcal{O}^i\}, \{\mathcal{A}^i\}, \mathcal{T}, \mathcal{R}, \gamma \rangle$, where $\mathcal{N} = \{1, \dots, N\}$ denotes the set of $N$ UAV agents.

At each timestep $t$, the true environment state $s_t \in \mathcal{S}$ is not directly accessible to any individual agent. Each agent $i$ receives a local observation $o_t^i \in \mathcal{O}^i$ derived from its onboard sensors, and selects an action $a_t^i \in \mathcal{A}^i$ according to its decentralised policy $\pi^i(a_t^i \mid o_{t-K:t}^i)$, conditioned on a fixed-length observation history of length $K$ to mitigate partial observability. The action spaces are continuous and identical across agents. Agents operate without explicit inter-agent communication; each policy relies solely on local observations.

Under the CTDE paradigm, the centralised critic accesses the global state $s_t$, constructed by concatenating all agents' local observations:
\begin{equation}
    s_t = \left[ o_t^1 \,\|\, o_t^2 \,\|\, \cdots \,\|\, o_t^N \right] \in \mathbb{R}^{N \cdot |\mathcal{O}|}
\end{equation}
whilst each decentralised actor conditions only on its own observation history during execution.

The state transition $s_{t+1} \sim \mathcal{T}(\cdot \mid s_t, \boldsymbol{a}_t)$ is governed by the joint action $\boldsymbol{a}_t = (a_t^1, \dots, a_t^N)$. The team receives a joint reward $r_t$ decomposed into a shared team term and individual agent terms, as detailed in Section~\ref{sec:reward}.

The objective is to find decentralised policies $\{\pi^i\}_{i=1}^{N}$ that maximise the expected discounted cumulative return:
\begin{equation}
    J(\{\pi^i\}) = \mathbb{E}\left[\sum_{t=0}^{T} \gamma^t r_t\right]
\end{equation}
where $\gamma \in (0,1)$ is the discount factor and $T$ is the episode horizon. The structural complexity of the environment, the sparsity of cooperative success signals, and the absence of inter-agent communication collectively render this optimisation problem non-trivial, motivating the mechanisms developed in the following sections.

\subsection{Overall Framework}

The proposed framework builds upon MASAC and operates under the CTDE paradigm. Each UAV maintains an independent mixture-of-experts actor (MoE Actor) network, which relies solely on its own local observation history during execution. A centralised critic accesses the global state, constructed by concatenating all agents' local observations during training, to provide global value estimates for each agent's policy update. No explicit inter-agent communication is assumed, and the architecture scales to an arbitrary number of agents.

Each agent's local observation vector has dimension 33, as detailed in Table~\ref{tab:obs}, comprising goal direction and distance, velocity, LiDAR sector readings, LiDAR frame difference, previous action, visitation memory, and the structural domain parameter $\omega$. To mitigate partial observability, the observation history is supplied to the actor as a fixed-length sequence of length $K$ via a sliding window. A shared LSTM encoder extracts temporal features from this sequence, after which a structure-aware gating network generates expert mixture weights conditioned on $\omega$, and the weighted outputs of $N$ sub-policy expert heads are fused to produce the final continuous action.

\begin{table}[t]
\centering
\caption{Composition of the local observation vector ($d_{\mathrm{obs}} = 33$)}
\label{tab:obs}
\renewcommand{\arraystretch}{1.2}
\begin{tabular}{clp{3.8cm}}
\toprule
\textbf{Obs. Index} & \textbf{Component} & \textbf{Description} \\
\midrule
1-3   & Goal vector            & Relative displacement to goal, normalised by $d_{\max}$ \\
4-6   & Velocity               & Agent velocity vector, normalised by $v_{\max}$ \\
7-14  & LiDAR readings         & Minimum range per sector, normalised to $[0,1]$ \\
15-22 & LiDAR frame delta      & Per-sector range difference between consecutive frames; positive values indicate an approaching obstacle \\
23-25 & Previous action        & Action executed at the preceding timestep \\
26-29 & Visitation memory      & Local stagnation features for the optima diagnosis module \\
30-33 & Domain parameter $\omega$ & Local geometric structural features for the generalisation module \\
\bottomrule
\end{tabular}
\end{table}

At the data management level, the framework maintains three categories of buffer. A standard experience replay buffer $\mathcal{R}$ stores all environment interaction trajectories. A cooperative success buffer $\mathcal{D}_{\mathrm{coop}}$ applies fine-grained per-scenario and per-start--goal-pair quota management, retaining only high-quality trajectories in which all agents successfully reach their respective goals. An individual demonstration buffer $\mathcal{D}^i$ is maintained independently for each agent and stores supplementary trajectories in which at least one agent reaches its goal. During each training update, the three buffers are sampled jointly with differentiated weights: the replay buffer provides environment exploration signal, whilst the demonstration buffers supervise the actor network directly via behavioural cloning.

\begin{figure}[t]
    \centering
    \includegraphics[width=\columnwidth]{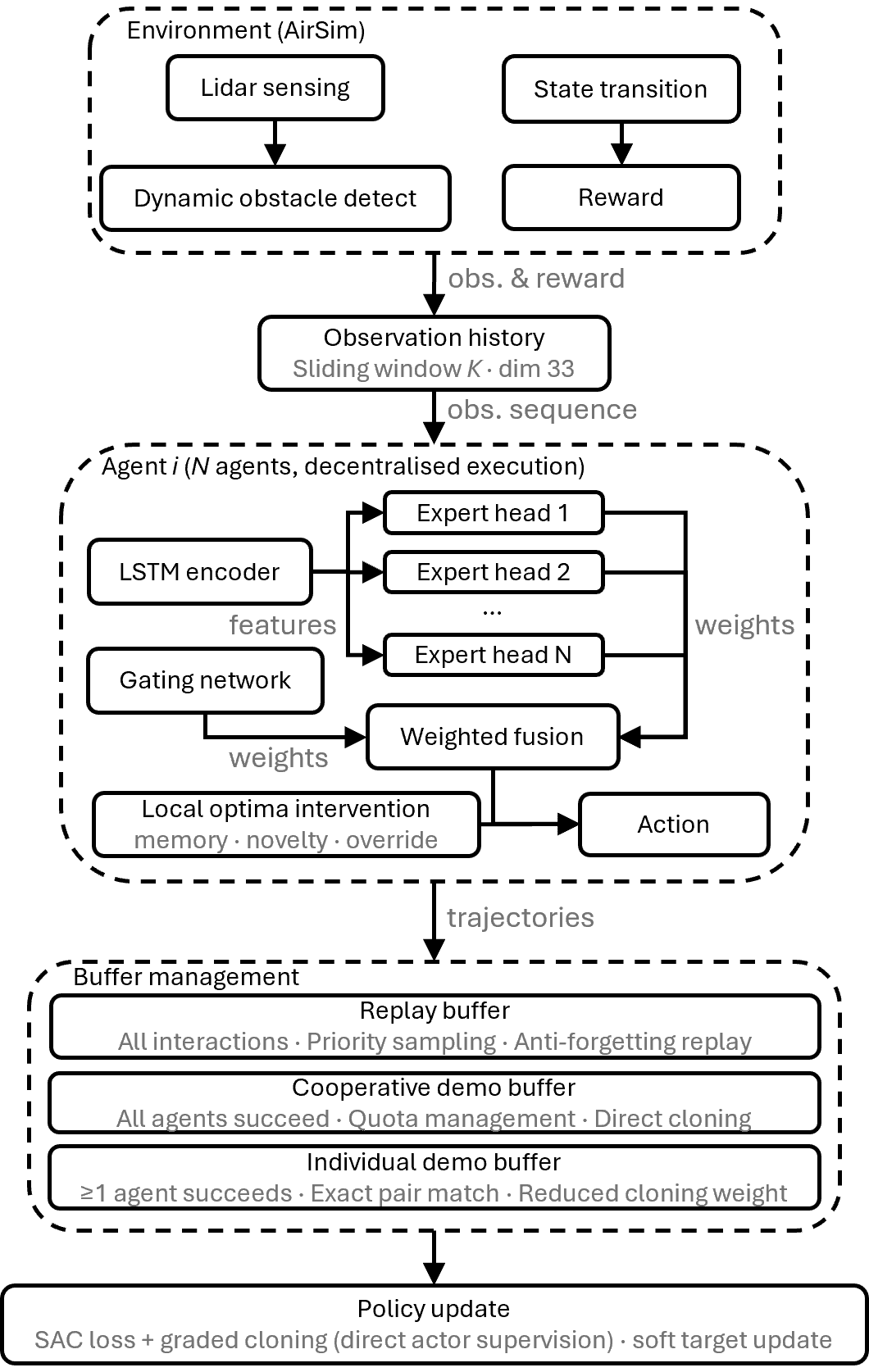}
    \caption{Overall architecture of the proposed cooperative multi-UAV navigation framework.}
    \label{fig:framework}
\end{figure}

A curriculum scheduler governs stage advancement using dual conditions on cooperative success rate and collision rate, and triggers retrospective evaluation and experience pre-filling over all previously mastered scenarios upon each stage transition. An execution-level local optima intervention module operates independently of the policy network without any additional trainable parameters, and directly overrides the policy output upon detection of agent stagnation. The overall framework architecture is illustrated in Fig.~\ref{fig:framework}.

\subsection{Reward Function Design}
\label{sec:reward}

The total reward assigned to agent $i$ at timestep $t$ decomposes into an individual shaping term and a shared team term:
\begin{equation}
    r_t = r_t^{\mathrm{team}} + \sum_{i=1}^{N} r_t^{i}
    \label{eq:reward_total}
\end{equation}

The individual reward $r_t^i$ accumulates the following terms:
\begin{equation}
\begin{aligned}
r_t^i \;=\;
&\underbrace{w_{d}\,\Delta d^{*} - w_{p}\,d_t}_{\text{progress}}
+\underbrace{w_{p}\,d_t\,\beta\,\kappa_t + w_{c}\,\kappa_t}_{\text{corridor compensation}} \\
&+\underbrace{\frac{w_{n}}{\sqrt{n_t+1}}}_{\text{novelty}}
-\underbrace{w_{\ell}\max(0,\,\rho_{0}-\rho_t)}_{\text{loop penalty}} \\
&-\underbrace{w_{s1}(\tau_1-d_{\ell})^{+} + w_{s2}(\tau_2-d_{\ell})^{+}}_{\text{proximity penalty}} \\
&+\underbrace{w_{\mathrm{align}}\,\alpha_t + w_{\mathrm{prog}}\,\Delta p_t\,\alpha_t}_{\text{goal alignment}}
-\underbrace{w_{\mathrm{opt}}\,\mathbb{I}[\mathrm{opt}]}_{\text{local optima}} \\
&-\underbrace{p_{\mathrm{col}}^i}_{\text{collision}}
+\underbrace{r_{\mathrm{arrive}}^i}_{\text{arrival}}
\end{aligned}
\label{eq:reward_ind}
\end{equation}

\noindent where $d_t$ is the current distance from agent $i$ to its goal; $\Delta d^{*} = d^{*}_{t-1} - d_t$ is the improvement over the best distance achieved so far; $\kappa_t \in [0,1]$ is a LiDAR-derived corridor confinement indicator, active only when the goal is occluded, which partially restores the distance penalty by a factor $\beta$ and provides a weak forward-motion signal to eliminate the reward vacuum during forced detours; $n_t$ is the visit count of the current cell; $\rho_t$ is the ratio of unique cells visited in a recent window, with $\rho_0$ as the target uniqueness threshold; $d_{\ell}$ is the minimum LiDAR range across all sectors, with $(x)^{+} = \max(0,x)$ and $\tau_1 > \tau_2$ as the two proximity warning thresholds; $\alpha_t = \cos\angle(\mathbf{v}_t, \hat{g}_t) \in [-1,1]$ is the cosine alignment between the executed velocity and the goal direction, active only when the goal is line-of-sight visible; $\Delta p_t$ is the normalised goal-progress increment; and $\mathbb{I}[\mathrm{opt}]$ is triggered by the local optima diagnosis module. The collision penalty $p_{\mathrm{col}}^i$ aggregates static obstacle, inter-agent, and dynamic obstacle collision terms. The arrival reward $r_{\mathrm{arrive}}^i$ is graded across three milestones: progressive, near-goal, and full arrival. All coefficients are tuneable hyperparameters.

The team reward $r_t^{\mathrm{team}}$ enforces cooperative synchronisation:
\begin{equation}
\begin{aligned}
r_t^{\mathrm{team}} \;=\;
&\underbrace{c_1\,s_t - c_2\,\delta_t}_{\text{synchronisation}}
-\underbrace{c_3\,\frac{\Delta T}{T}}_{\text{arrival gap}}
+\underbrace{b_{\mathrm{event}}}_{\text{cooperative events}}
\end{aligned}
\label{eq:reward_team}
\end{equation}

\noindent where $s_t = \min_i p_t^i$, with $p_t^i \in [0,1]$ denoting the normalised goal-progress of agent $i$; $s_t$ ensures the synchronisation bonus activates only when all agents are advancing concurrently; $\delta_t$ is the inter-agent distance gap; $\Delta T$ is the difference in arrival timesteps between agents, normalised by the episode length $T$, penalising asynchronous arrivals in proportion to the temporal disparity; and $b_{\mathrm{event}}$ is a sparse event bonus comprising all-agent arrival reward, all-agent near-goal reward, all-agent progressive reward, and a team collision penalty. All coefficients $c_1$, $c_2$, $c_3$, and the magnitudes of $b_{\mathrm{event}}$ are tuneable hyperparameters.

\subsection{Local Optima Diagnosis and Execution-Level Intervention}

The mechanism comprises three sub-modules, namely visited-state memory, local optima diagnosis, and execution-level intervention, operating in concert without introducing any additional parameters.

The environment is discretised into a uniform grid, and an independent visitation frequency table $\{c(\mathbf{x})\}$ is maintained for each agent, recording the cumulative visit count of each cell. At every timestep, the agent maps its current position to the corresponding cell and updates the count. The memory features are additionally encoded into a four-dimensional vector and appended to the observation (indices 26-29 in Table~\ref{tab:obs}), enabling the policy to perceive its own historical behavioural patterns directly.

At each timestep, a diagnosis is performed over a recent history window of length $W$ by evaluating four conditions:
\begin{equation}
\mathbb{I}[\mathrm{opt}] = \mathbb{I}\!\left[\mathcal{C}_{\mathrm{hover}} \vee \mathcal{C}_{\mathrm{repeat}} \vee \mathcal{C}_{\mathrm{block}} \vee \mathcal{C}_{\mathrm{wander}}\right]
\label{eq:local_opt}
\end{equation}
where $\mathcal{C}_{\mathrm{hover}}$ holds when the spatial span of the recent trajectory falls below a threshold, indicating in-place hovering; $\mathcal{C}_{\mathrm{repeat}}$ holds when the recent unique-cell ratio $\rho_t$ drops below the target threshold $\rho_0$, indicating repeated traversal of the same region; $\mathcal{C}_{\mathrm{block}}$ holds when the agent is close to the goal yet exhibits near-zero mean progress alongside a low mean LiDAR range, indicating obstruction by a nearby structure; and $\mathcal{C}_{\mathrm{wander}}$ holds when the agent is far from the goal, shows near-zero mean progress, but maintains a large trajectory span, indicating large-scale wandering around obstacles. When $\mathbb{I}[\mathrm{opt}] = 1$ and the agent is more than a safety distance $d_{\min}$ from its goal, the penalty $w_{\mathrm{opt}}\,\mathbb{I}[\mathrm{opt}]$ is applied to the individual reward and the execution-level intervention timer is activated.

Whilst the intervention timer is active, the policy output is directly overridden and the agent is steered towards an escape direction selected by a directional novelty scoring function. For each candidate direction $\phi_k$, the score is computed as:
\begin{equation}
f(\phi_k) = w_{\nu}\,\nu(\phi_k) + w_{\mathrm{cl}}\,\mathrm{cl}(\phi_k) - w_{r}\,\mathrm{pen}(\phi_k)
\label{eq:escape_score}
\end{equation}
where $\nu(\phi_k)$ is the directional novelty, estimated by probing cells along the candidate direction and aggregating inverse square-root visit counts; $\mathrm{cl}(\phi_k)$ is the forward clearance derived from LiDAR readings; and $\mathrm{pen}(\phi_k)$ aggregates backtracking and revisit penalties. When a dead-end is detected, defined as the number of blocked candidate directions exceeding a threshold, the scoring function switches to a clearance-dominant mode, prioritising the most open direction over the most novel one. The intervention mechanism serves as an active guide during the early stages of training, steering agents away from stagnation and generating informative transitions that enrich the replay buffer. As training progresses, the policy progressively internalises the ability to avoid local optima through the accumulated reward signal and the visitation memory encoded in the observation, reducing its reliance on the intervention mechanism accordingly.

\subsection{Hierarchical Demo Buffer with Graded Cloning}

In cooperative navigation tasks, successful events in which all agents reach their respective goals are extremely sparse during the early stages of training, resulting in a lack of effective cooperative learning signals. Existing demonstration learning methods~\cite{argall2009survey,ho2016generative} treat all demonstration data equally, failing to distinguish between fully cooperative completion and the success of only some agents, thereby diluting the supervisory signal from high-quality cooperative trajectories with lower-quality samples. Furthermore, existing methods typically rely on independent auxiliary policy networks to distil demonstrations and transfer them to the main actor, introducing cumulative errors from two-stage approximation. To address this, a hierarchical demonstration buffer and a graded behavioural cloning mechanism are proposed, treating the degree of cooperative completion as the primary attribute of demonstration data, thereby mitigating learning imbalance among agents at both the data-management and policy-optimisation levels.

Two independent buffers are maintained, each storing demonstration data of a different quality level. The cooperative success buffer, with total capacity $M_{\mathrm{coop}}$, stores complete trajectories in which all agents successfully reach their respective goals. The individual success buffer, with capacity $M_{\mathrm{ind}}$ per agent, stores trajectories in which at least one agent succeeds, serving as a supplementary source during early training when fully cooperative samples are scarce. The cooperative success buffer employs a two-tiered quota management scheme at both the scenario level and the start--goal pair level, with per-pair capacity $M_{\mathrm{pair}}$ and per-scenario capacity $M_{\mathrm{maze}}$. When a quota is reached, a quality-based replacement strategy removes trajectories with lower cooperative scores, ensuring that high-quality demonstrations across scenarios of varying difficulty are retained in a balanced manner. Scenario and start--goal pair identifiers are used as explicit dimensions for buffer management, preventing successful samples from simpler scenarios in later training from overshadowing effective demonstrations from earlier, more challenging ones.

At each actor gradient update, samples are drawn from both buffers and behavioural cloning supervision is applied directly to the actor network using hierarchical weights:
\begin{align}
\mathcal{L}_{\mathrm{BC}}^{i} &=
\mathbb{E}_{(o,\,a^{*})\sim\mathcal{D}_{\mathrm{coop}}}
\!\left[\bigl\|\pi_{i}(o)-a^{*}\bigr\|_{2}^{2}\right] \nonumber \\
&+\,\lambda\,
\mathbb{E}_{(o,\,a^{*})\sim\mathcal{D}^i}
\!\left[\bigl\|\pi_{i}(o)-a^{*}\bigr\|_{2}^{2}\right]
\label{eq:bc_loss}
\end{align}
where $\mathcal{D}_{\mathrm{coop}}$ denotes the cooperative success buffer, $\mathcal{D}^i$ denotes the individual success buffer for agent $i$, and $\lambda < 1$ is a tuneable weighting coefficient reflecting the lower cooperative quality of partial success demonstrations relative to fully cooperative ones. The effective cloning coefficient $\beta_{\mathrm{coop}}$ is adapted dynamically throughout training, increasing linearly from an initial value $\beta_{\min}$ to a maximum $\beta_{\max}$ in proportion to the occupancy of $\mathcal{D}_{\mathrm{coop}}$. This schedule introduces a modest cloning signal during early training to avoid over-constraining exploration, whilst progressively strengthening the supervision as cooperative demonstrations accumulate. The cloning supervision is applied directly to the actor network without routing through a separately trained auxiliary policy network as an intermediary, thereby eliminating the cumulative approximation error inherent in two-stage policy distillation and improving demonstration exploitation efficiency under sparse cooperative success signals. The total actor loss is the sum of the SAC policy gradient loss and the behavioural cloning loss:
\begin{equation}
\mathcal{L}_{\mathrm{actor}}^i = \mathcal{L}_{\mathrm{SAC}}^i + \beta_{\mathrm{coop}}\,\mathcal{L}_{\mathrm{BC}}^i
\label{eq:actor_loss}
\end{equation}

The centralised critic is updated by minimising the difference between the current value estimate and the bootstrapped target value computed from replay buffer samples, which include successful trajectories stored alongside standard interactions. Through this mechanism, the critic implicitly learns to assign higher value estimates to states and actions along successful cooperative paths, complementing the explicit behavioural cloning supervision applied to the actor.

\subsection{Safety-aware Dual-condition Curriculum Scheduling}

The training process is organised as a sequence of progressively more demanding scenario stages. Unlike existing curriculum learning methods that advance stages based solely on task success rate, the proposed mechanism requires both the all-agent cooperative success rate $\eta$ and the collision rate $\xi$ to jointly satisfy $\eta \geq \tau_{\eta}$ and $\xi \leq \tau_{\xi}$ before advancement is permitted, where $\tau_{\eta}$ and $\tau_{\xi}$ are independently configurable per scenario. When the success rate criterion is met but the collision rate remains above its ceiling, stage advancement is blocked, compelling the policy to reduce collision risk whilst sustaining task performance, thereby enforcing concurrent guarantees on both objectives and preventing the policy from exploiting high-risk behaviours to achieve superficially favourable success metrics.

Upon each stage advancement, a retrospective evaluation is immediately conducted over all previously mastered scenarios. Any scenario falling below its threshold is reinstated into the training queue, and fresh trajectories are collected from all passed scenarios at a rate of $K_{\mathrm{prefill}}$ episodes per scenario and injected into the replay buffer, ensuring representative historical data is retained following each transition. During normal training, a fraction $p_{\mathrm{hist}}$ of episodes are drawn from previously mastered scenarios to maintain continual reinforcement of acquired behaviours alongside the learning of new ones. Together with the retrospective evaluation and experience pre-filling, this constitutes a principled countermeasure against catastrophic forgetting across training scenarios.

\subsection{Structure-aware Generalisation Mechanism}

Complex constrained environments can be characterised by a finite repertoire of recurring local geometric patterns that exhibit similar sensor-level feature distributions across spatially distinct scenarios. Existing generalisation methods target variations in physical dynamics or task objectives rather than local geometric repeatability, causing policies to implicitly memorise coordinate-specific information from training scenarios. The proposed mechanism addresses this by abstracting recurring local geometric patterns into a domain parameter computed deterministically from real-time onboard sensor readings, enabling structure-aware cross-scenario transfer without prior knowledge of the target environment.

The components of the domain parameter $\omega$ should be selected as local geometric features that recur within the target environment type, are computable in real time from onboard sensors, and are mutually distinguishable from one another. The specific choice of features depends on the structural characteristics of the deployment scenario. In the maze test scenarios used in this work, the most representative local structural variations can be categorised into four types: dead-end structures, single-wall structures characterised by pronounced lateral occlusion differences, narrow entrance structures with only a small number of consecutive clear directions, and open-area structures in which the majority of directions remain clear. These four structural types are mutually distinguishable at the LiDAR-reading level and collectively capture the predominant local geometric variations in maze environments. They are therefore selected as the domain parameter components for validating the proposed mechanism.

At each timestep, the domain parameter $\omega \in [0,1]^4$ is computed deterministically from the 8-sector normalised LiDAR readings of the current frame, requiring neither manual annotation nor any prior knowledge of the current scenario. The four components $\omega_{\mathrm{dead}}$, $\omega_{\mathrm{wall}}$, $\omega_{\mathrm{narrow}}$, and $\omega_{\mathrm{open}}$ quantify the degree of dead-end structure, lateral occlusion asymmetry, narrow-entrance constriction, and overall openness, respectively, corresponding to the four local structural types identified above. Let $b_k \in \{0,1\}$ denote whether the $k$-th LiDAR sector is occluded by an obstacle, determined by whether the normalised LiDAR reading falls below a threshold $\theta$; $n_{\mathrm{block}} = \sum_k b_k$ denotes the number of occluded sectors; and $g$ denotes the maximum number of consecutive clear sectors. The four components are defined as:
\begin{equation}
\left\{
\begin{array}{@{}l@{}}
\omega_{\mathrm{dead}} =
\mathrm{clip}\!\left(\tfrac{n_{\mathrm{block}}-5}{3},0,1\right)
\cdot
\mathrm{clip}\!\left(1-\tfrac{g-1}{2},0,1\right) \\[0.25em]
\omega_{\mathrm{wall}} =
\left|\bar{b}_{\mathrm{left}}-\bar{b}_{\mathrm{right}}\right| \\[0.25em]
\omega_{\mathrm{narrow}} =
\mathrm{clip}\!\left(\tfrac{\min(g,3)-1}{2},0,1\right)
\cdot
\mathrm{clip}\!\left(\tfrac{n_{\mathrm{block}}}{5},0,1\right) \\[0.25em]
\omega_{\mathrm{open}} =
\frac{8-n_{\mathrm{block}}}{8}
\end{array}
\right.
\label{eq:omega_components}
\end{equation}
where $\bar{b}_{\mathrm{left}}$ and $\bar{b}_{\mathrm{right}}$ denote the mean occlusion values of the left and right four LiDAR sectors, respectively. The component $\omega_{\mathrm{dead}}$ approaches unity when most sectors are occluded and the maximum consecutive clear gap is minimal, and is used to represent dead-end structures. The component $\omega_{\mathrm{wall}}$ quantifies lateral occlusion asymmetry between the two sides; it takes a high value when one side is dominated by a wall or obstacle whilst the other remains relatively clear, and is therefore used to represent single-wall structures. The component $\omega_{\mathrm{narrow}}$ takes a high value when only two to three consecutive clear sectors exist amid heavy surrounding occlusion, and is used to represent narrow entrances or narrow passages. The component $\omega_{\mathrm{open}}$ reflects the overall proportion of clear sectors among all LiDAR sectors, and is used to represent open areas. These four components are designed with explicit structural meanings and together provide a compact low-dimensional representation of the principal local geometric variations in constrained environments. The domain parameter $\omega$ is additionally appended as the final four dimensions of the observation vector, corresponding to indices 30-33 in Table~\ref{tab:obs}, enabling the policy to directly perceive the current local geometric structure at the observation level.

The actor network adopts a mixture-of-experts architecture comprising a shared LSTM encoder, a structure-aware gating network, and $N_e$ expert heads. The shared encoder processes the observation history $o_{t-K:t}^i$ of length $K$ through a linear projection layer, an LSTM, and a fully connected layer to produce a temporal feature representation $\mathbf{h}$:
\begin{equation}
\mathbf{h} = \mathrm{MLP}\bigl(\mathrm{LSTM}(\mathrm{proj}(o_{t-K:t}^i))\bigr)
\end{equation}
The structure-aware gating network takes $\omega$ as input and produces a mixture weight vector over the $N_e$ expert heads via a two-layer fully connected network followed by softmax normalisation:
\begin{equation}
\mathbf{g} = \mathrm{softmax}\bigl(\mathrm{MLP}_{\mathrm{gate}}(\omega)\bigr) \in \Delta^{N_e - 1}
\end{equation}
where $\Delta^{N_e-1}$ denotes the $N_e$-dimensional probability simplex. Each of the $N_e$ expert heads independently takes the shared feature $\mathbf{h}$ as input and produces its own action mean $\mu_k$ and log standard deviation $\log\sigma_k$ via independent linear projections, for $k = 1,\ldots,N_e$. The final action distribution mean and log standard deviation are obtained by weighted fusion of all expert outputs according to the gate weights:
\begin{equation}
\left\{
\begin{array}{@{}l@{}}
\mu = \displaystyle\sum_{k=1}^{N_e} g_k\,\mu_k \\[0.6em]
\log\sigma = \mathrm{clip}\!\left(\displaystyle\sum_{k=1}^{N_e} g_k\,\log\sigma_k,\;\sigma_{\min},\;\sigma_{\max}\right)
\end{array}
\right.
\label{eq:moe_fusion}
\end{equation}
where $\sigma_{\min}$ and $\sigma_{\max}$ are the lower and upper bounds applied to the log standard deviation to ensure numerical stability. The action is subsequently squashed to $[-1,1]^3$ via a tanh transformation. The entire forward computation path is differentiable, and the gating network and expert heads are trained end-to-end through the SAC policy gradient without requiring any expert labels or structural type supervision.

Since $\omega$ is computed deterministically from sensor readings and is entirely decoupled from scenario-specific coordinates, the policy learns behavioural responses conditioned on local geometric features rather than memorising the spatial layouts of particular training scenarios. During training, the gating network learns to dynamically adjust the activation weights of different expert sub-policies according to the current local structure, whilst each expert head progressively specialises in responding to different local structural patterns. Consequently, any new scenario composed of structural patterns encountered during training can be deployed directly without retraining, achieving cross-scenario transfer. The mechanism is also scalable: the computation of $\omega$ relies solely on onboard sensors, the number of expert heads $N_e$ can be adjusted according to environmental complexity, and the overall framework does not depend on a specific scenario type or number of agents.

\subsection{Training Procedure}

The training procedure integrates all previously described mechanisms into a unified learning loop. Prior to training, the framework conducts a startup evaluation over all previously mastered scenarios, registers those that pass validation into the history replay queue, and pre-fills the replay buffer with trajectories collected from these scenarios, ensuring that the buffer contains representative data from mastered scenarios before training begins.

At each timestep, each agent samples an action from its current policy (during the warm-up phase of $T_{\mathrm{warm}}$ steps, uniformly random actions are used to initialise the buffer). The resulting transition tuple is stored in the replay buffer with the current scenario identifier and start--goal pair identifier attached to support priority sampling. The local optima intervention module operates independently at each step and directly overrides the policy output whenever the diagnosis condition $\mathbb{I}[\mathrm{opt}] = 1$ is triggered.

At the end of each episode, the demonstration buffers are updated according to the outcome. If all agents reach their respective goals, the trajectory undergoes quality filtering based on the inter-agent arrival gap; trajectories satisfying the gap threshold are stored into the cooperative success buffer, with higher-quality trajectories stored multiple times. If exactly one agent reaches its goal and the episode belongs to the current curriculum stage scenario and its corresponding start--goal pair, that agent's trajectory is stored into its individual success buffer. Trajectories from other scenarios or start--goal pairs are excluded to prevent behaviourally inconsistent demonstrations from contaminating the cloning supervision. Let $\tau_{\mathrm{gap}}$ denote the maximum permissible inter-agent arrival gap, in timesteps, for a trajectory to qualify for storage in $\mathcal{D}_{\mathrm{coop}}$; trajectories with arrival gap exceeding $\tau_{\mathrm{gap}}$ are discarded.

After an initial collection phase of $T_{\mathrm{after}}$ steps to populate the replay buffer, every $T_{\mathrm{update}}$ steps a policy update is performed by jointly sampling from the replay buffer, the cooperative success buffer, and the individual success buffers. The SAC policy gradient loss and the graded behavioural cloning loss are combined as $\mathcal{L}_{\mathrm{actor}}^i = \mathcal{L}_{\mathrm{SAC}}^i + \beta_{\mathrm{coop}}\,\mathcal{L}_{\mathrm{BC}}^i$, and gradient updates are applied separately to each actor and the centralised critic. The cloning coefficient $\beta_{\mathrm{coop}}$ increases dynamically with the occupancy of the cooperative success buffer, remaining small during early training to avoid over-constraining exploration.

Every $T_{\mathrm{eval}}$ episodes, a dedicated evaluation is conducted on the current curriculum stage, computing the all-agent cooperative success rate $\eta$ and the collision rate $\xi$. If the dual condition $\eta \geq \tau_{\eta}$ and $\xi \leq \tau_{\xi}$ is jointly satisfied, retrospective evaluation and experience pre-filling are triggered over all previously mastered scenarios, after which training advances to the next curriculum stage. During normal training, each episode is assigned to a previously mastered scenario with probability $p_{\mathrm{hist}}$ and to the current stage with probability $1 - p_{\mathrm{hist}}$. The complete training procedure is summarised in Algorithm~\ref{alg:training}.

\begin{algorithm*}[t]
\SetAlgoLined
\DontPrintSemicolon
\caption{Training procedure for cooperative multi-UAV navigation}
\label{alg:training}
\SetKwInOut{Input}{Input}
\Input{Scenarios $\{s_1,\ldots,s_S\}$ in curriculum order, thresholds $\{\tau_{\eta}^s, \tau_{\xi}^s\}$; actors $\{\pi^i\}_{i=1}^N$, critic $Q$}
Initialise $\mathcal{R}$, $\mathcal{D}_{\mathrm{coop}}$, $\{\mathcal{D}^i\}_{i=1}^N$; mastered set $\mathcal{M} \leftarrow \emptyset$; pre-fill $\mathcal{R}$ from any stages passing startup evaluation\;
\For{each environment timestep $t$}{
    $a^i \leftarrow \mathrm{Uniform}(\mathcal{A}^i)$ if $t \leq T_{\mathrm{warm}}$, else $\pi^i(o_{t-K:t}^i)$ with local optima override if $\mathbb{I}[\mathrm{opt}]=1$\;
    Execute $\{a^i\}$; store transition in $\mathcal{R}$ tagged with scenario and start--goal pair identifiers\;
    \If{episode terminates}{
        \lIf{all agents arrived and gap $\leq \tau_{\mathrm{gap}}$}{store trajectory $\rightarrow \mathcal{D}_{\mathrm{coop}}$}
        \lElseIf{one agent arrived and same stage and pair}{store that agent's trajectory $\rightarrow \mathcal{D}^i$}
        Select next episode from $\mathcal{M}$ w.p.\ $p_{\mathrm{hist}}$, else from current stage\;
        \If{episode count $\bmod\, T_{\mathrm{eval}} = 0$}{
            Evaluate current stage $\rightarrow \eta, \xi$\;
            \If{$\eta \geq \tau_{\eta}$ and $\xi \leq \tau_{\xi}$}{
                Retrospective evaluation and experience pre-fill over $\mathcal{M}$; add current stage to $\mathcal{M}$; advance stage\;
            }
        }
    }
    \If{$t \geq T_{\mathrm{after}}$ and $t \bmod T_{\mathrm{update}} = 0$}{
        Sample minibatches from $\mathcal{R}$, $\mathcal{D}_{\mathrm{coop}}$, $\{\mathcal{D}^i\}$; update $\{\pi^i\}$ and $Q$ via $\mathcal{L}_{\mathrm{actor}}^i = \mathcal{L}_{\mathrm{SAC}}^i + \beta_{\mathrm{coop}}\mathcal{L}_{\mathrm{BC}}^i$; adapt $\beta_{\mathrm{coop}}$\;
    }
}
\end{algorithm*}

\section{Simulation Setup}

Experiments are conducted using the Microsoft AirSim simulation platform~\cite{shah2017airsim}, with Unreal Engine 4 as the 3D rendering and physics engine. Each UAV is equipped with a single-channel, horizontally omnidirectional LiDAR sensor with a range of 12\,m, a rotation rate of 10\,r/s, and a point cloud density of 5000 points/s, mounted 0.1\,m above the vehicle body. Raw LiDAR point clouds are aggregated into 8 equal angular sectors, with the minimum range per sector extracted and normalised to $[0,1]$ as the perceptual input. Each UAV takes off and hovers at an altitude of 1.8\,m, with a 3D continuous velocity command $\mathbf{a}^i \in [-1,1]^3$ as the action space, corresponding to forward/backward, lateral, and vertical velocity components respectively. All training and evaluation are conducted on a workstation equipped with an NVIDIA RTX 6000 Ada Generation GPU (49\,GB VRAM) under CUDA 12.8.

\begin{figure}[t]     
\centering     
\includegraphics[width=\columnwidth]{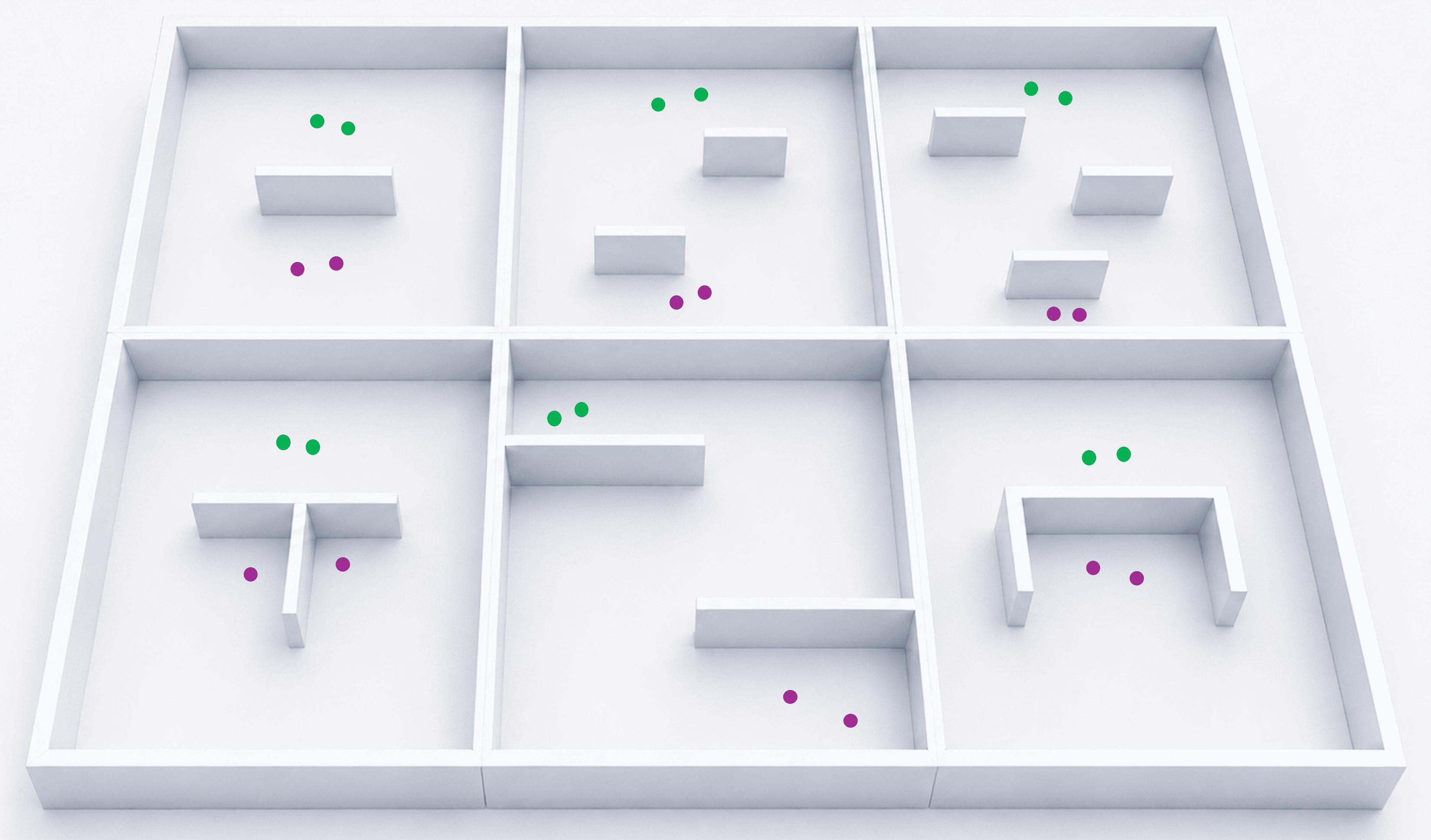}     \caption{Overview of training scenarios maze\_01 to maze\_06 (20\,m\,$\times$\,20\,m each), numbered clockwise from the top-left. Purple and green markers indicate start and goal positions respectively. Each scenario is configured with multiple start--goal pairs; during training, start and goal positions are swapped and agent assignments are exchanged across pairs to promote generalisation and prevent overfitting to specific spatial configurations.}     \label{fig:maze_1_6} \end{figure}

\begin{figure}[t]     
\centering     
\includegraphics[width=0.5\columnwidth]{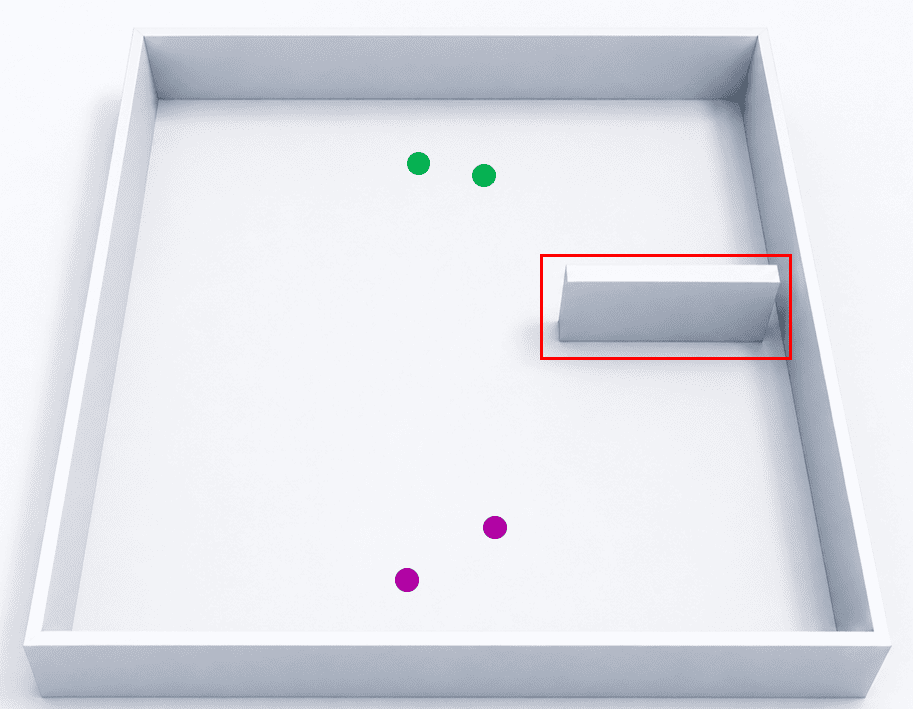}     \caption{Training scenario maze\_07 with dynamic obstacles. The obstacle enclosed within the red bounding box is a moving obstacle performing lateral reciprocating motion at 1.5\,m/s within a bounded region. Purple and green markers indicate start and goal positions respectively.}     \label{fig:maze_7} \end{figure}

Training uses 7 maze scenarios constructed as enclosed 3D environments in Unreal Engine 4. Scenarios maze\_01 to maze\_06 occupy a 20\,m\,$\times$\,20\,m area, whilst scenario maze\_07 is constructed within a larger 60\,m\,$\times$\,40\,m environment to accommodate the lateral reciprocating motion of the moving obstacles. Each scenario is configured with multiple start--goal pairs, encompassing forward, reverse, and agent-swapped navigation paths, ensuring that the policy is trained across diverse spatial configurations within the same scenario and does not overfit to specific start--goal coordinates. Scenarios maze\_01 to maze\_03 are basic obstacle scenarios comprising wall structures and static obstacles. Scenarios maze\_04 to maze\_06 introduce more complex obstacle layouts with pronounced local optima characteristics, requiring the policy to develop active stagnation-escape capabilities. Scenario maze\_07 further incorporates moving obstacles to validate the framework's adaptability in dynamic environments. The training scenario layouts are illustrated in Fig.~\ref{fig:maze_1_6} and Fig.~\ref{fig:maze_7}. A separate test scenario, unseen during training, is reserved exclusively for evaluating cross-scenario transfer.  

  

Four metrics are used for quantitative evaluation: the all-agent cooperative success rate $\eta$, defined as the proportion of episodes in which all agents reach their respective goals within the maximum episode length $T$; the collision rate $\xi$, defined as the proportion of episodes in which any collision event occurs; the mean terminal distance, defined as the average distance from each agent to its goal at episode termination; and the inter-agent arrival gap, defined as the difference in arrival timesteps between agents in fully successful episodes, measuring cooperative synchronisation. In generalisation experiments, the gate weight distribution $\mathbf{g}$ is additionally recorded to analyse the activation patterns of the structure-aware mechanism across different local geometric structures. All hyperparameter settings are listed in Table~\ref{tab:hyperparams}.

\begin{table}[t]
\centering
\caption{Hyperparameter configuration.}
\label{tab:hyperparams}
\renewcommand{\arraystretch}{1.15}
\setlength{\tabcolsep}{4pt}
\begin{tabular}{p{3.8cm}ll}
\toprule
\textbf{Parameter} & \textbf{Symbol} & \textbf{Value} \\
\midrule
\multicolumn{3}{l}{\textit{Network architecture}} \\
Observation history length & $K$ & 16 \\
Hidden layer dimension & — & 128 \\
Number of expert heads & $N_e$ & 4 \\
Structural parameter dim. & — & 4 \\
\midrule
\multicolumn{3}{l}{\textit{Training configuration}} \\
Warm-up steps & $T_{\mathrm{warm}}$ & 3{,}000 \\
Policy update interval & $T_{\mathrm{update}}$ & 50 \\
Minibatch size & — & 128 \\
Discount factor & $\gamma$ & 0.99 \\
Soft target update coeff. & — & 0.001 \\
Actor learning rate & — & $5\times10^{-5}$ \\
Critic learning rate & — & $5\times10^{-5}$ \\
Entropy coeff. learning rate & — & $2\times10^{-5}$ \\
\midrule
\multicolumn{3}{l}{\textit{Replay and demo buffers}} \\
Replay buffer capacity & — & 120{,}000 \\
Coop. demo buffer capacity & $M_{\mathrm{coop}}$ & 60{,}000 \\
Indiv. demo buffer capacity & $M_{\mathrm{ind}}$ & 5{,}000 \\
\midrule
\multicolumn{3}{l}{\textit{Behavioural cloning}} \\
Initial cloning coefficient & $\beta_{\min}$ & 0.03 \\
Maximum cloning coefficient & $\beta_{\max}$ & 0.40 \\
Indiv. demo weight ratio & $\lambda$ & 1/3 \\
Arrival gap threshold & $\tau_{\mathrm{gap}}$ & 100 steps \\
\midrule
\multicolumn{3}{l}{\textit{Curriculum scheduling}} \\
Evaluation interval & $T_{\mathrm{eval}}$ & 100 episodes \\
History mixing ratio & $p_{\mathrm{hist}}$ & 0.40 \\
Pre-fill episodes per scenario & $K_{\mathrm{prefill}}$ & 15 \\
\bottomrule
\end{tabular}
\end{table}

\section{Results and Analysis}

\subsection{Overall Performance}

Table~\ref{tab:main_results} reports the evaluation results of the proposed framework across all training scenarios and the unseen test scenario, each evaluated over 20 episodes. The framework surpasses the corresponding curriculum advancement threshold $\tau_{\eta}$ on all six static maze scenarios (maze\_01 to maze\_06), successfully completing all curriculum stages. Scenarios maze\_02 and maze\_06 achieve a cooperative success rate of 1.000, and maze\_01 achieves 0.950, demonstrating highly reliable cooperative navigation in structurally regular environments. Scenario maze\_04, which features complex obstacle layouts with pronounced local optima characteristics, yields a lower success rate (0.800) and a higher collision rate (0.400), reflecting the greater demand placed on stagnation-escape capability; nonetheless, the framework still meets the per-scenario thresholds ($\tau_{\eta}=0.75$, $\tau_{\xi}=0.40$), validating the safety-aware dual-condition curriculum scheduling mechanism.

\begin{table}[t]
\centering
\caption{Evaluation results of the proposed framework across all scenarios (20 episodes each). $\eta$: cooperative success rate; $\xi$: collision rate; Dist.: mean terminal distance.}
\label{tab:main_results}
\renewcommand{\arraystretch}{1.15}
\begin{tabular}{llccc}
\toprule
\textbf{Scenario} & \textbf{Type} & $\eta$ & $\xi$ & \textbf{Dist.\ (m)} \\
\midrule
maze\_01 & Basic static    & 0.950 & 0.050 & 1.783 \\
maze\_02 & Basic static    & 1.000 & 0.000 & 1.539 \\
maze\_03 & Basic static    & 0.900 & 0.150 & 2.083 \\
maze\_04 & Local optima    & 0.800 & 0.400 & 3.348 \\
maze\_05 & Local optima    & 0.800 & 0.050 & 2.510 \\
maze\_06 & Local optima    & 1.000 & 0.000 & 1.527 \\
maze\_07 & Dynamic obs.    & 0.800 & 0.200 & 3.474 \\
\midrule
maze\_mix & Unseen (test)  & 0.750 & 0.100 & 3.260 \\
\bottomrule
\end{tabular}
\end{table}

Scenario maze\_07 contains three moving obstacles, each performing lateral reciprocating motion within a bounded region. The framework achieves a success rate of 0.800 and a collision rate of 0.200, with a mean terminal distance of 3.474\,m. Dynamic obstacles are detected in all episodes, demonstrating the framework's adaptability in dynamic environments.

\begin{figure}[t]
    \centering
    \includegraphics[width=\columnwidth]{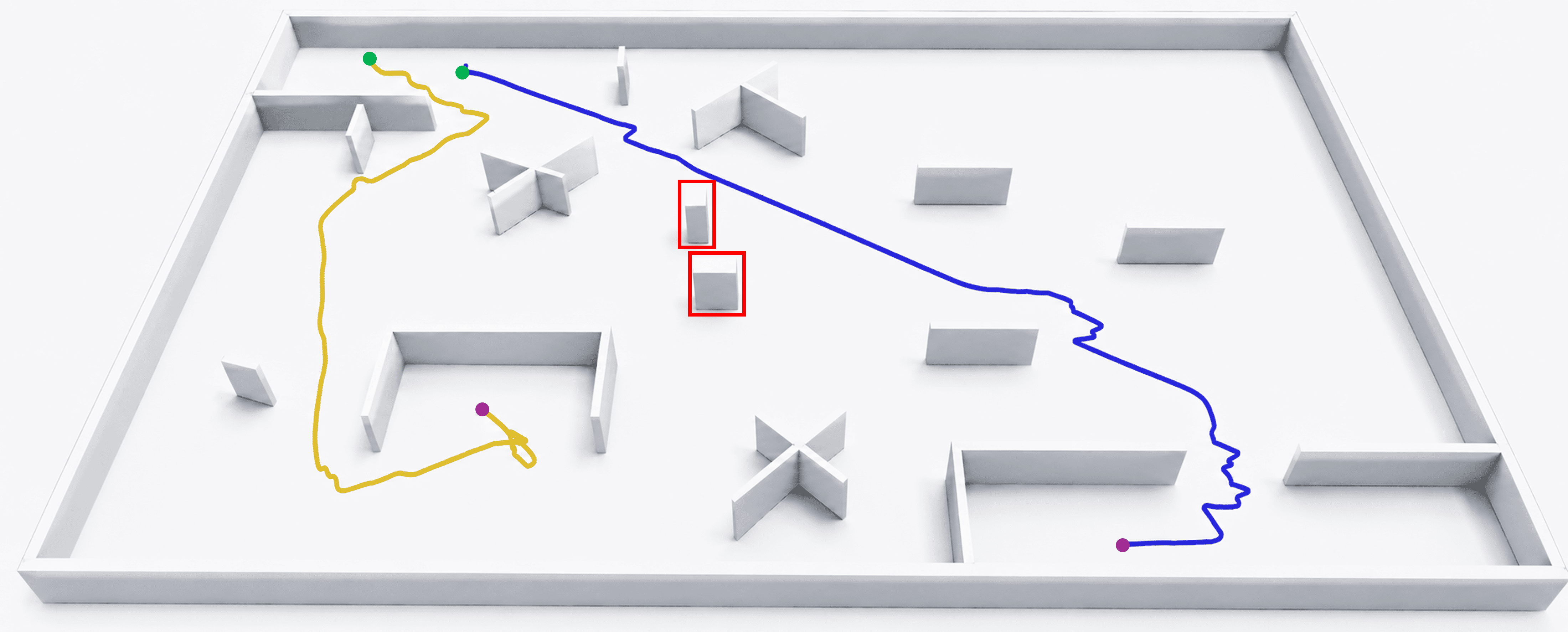}
    \caption{Representative UAV flight trajectories in the unseen test scenario maze\_mix. Yellow and blue curves denote the trajectories of the two UAVs respectively; purple markers indicate start positions, green markers indicate goal positions, and obstacles enclosed within red bounding boxes are moving obstacles.}
    \label{fig:traj}
\end{figure}

On the unseen test scenario maze\_mix, the framework achieves a cooperative success rate of 0.750 under dynamic obstacle conditions across all episodes, with a mean collision rate of 0.100 and a mean terminal distance of 3.260\,m. Fig.~\ref{fig:traj} illustrates the 3D flight trajectories of the two UAVs during a representative successful episode in maze\_mix. The yellow and blue curves correspond to the two UAVs respectively, with purple markers indicating start positions and green markers indicating goal positions. Both UAVs autonomously navigate around static and moving obstacles performing lateral reciprocating motion, and reach their respective goals without any prior knowledge of the scenario, validating the cross-scenario transfer capability of the proposed framework.

\subsection{Comparison with Baseline Methods}

To evaluate the advantage of the proposed framework over existing approaches, experiments are conducted on maze\_05 and the unseen test scenario maze\_mix against two baseline methods: MAPPO~\cite{yu2022surprising} and Standard MASAC, a variant of the proposed framework with all proposed mechanisms removed. All methods are trained for 1500 episodes under identical environment and collision detection conditions to ensure a fair comparison. It should be noted that MAPPO operates on a 21-dimensional observation space without the structural domain parameter $\omega$, whilst the full framework uses a 33-dimensional observation which incorporates $\omega$ as an additional input to the structure-aware gating network. This difference in observation design is an inherent part of the proposed framework rather than an experimental inconsistency, as the construction of $\omega$ from real-time LiDAR readings constitutes one of the core contributions of this work. Results are summarised in Table~\ref{tab:comparison}.

On the training scenario maze\_05, the full framework achieves a cooperative success rate of 0.800, outperforming MAPPO (0.550) and Standard MASAC (0.500). The collision rate of the full framework (0.050) is substantially lower than that of MAPPO (0.200) and Standard MASAC (0.400), validating the effectiveness of the safety-aware dual-condition curriculum scheduling mechanism in concurrently guaranteeing task performance and flight safety.

On the unseen test scenario maze\_mix, the full framework achieves a zero-shot cooperative success rate of 0.750 with a mean terminal distance of 3.260\,m, significantly outperforming both baselines. MAPPO suffers a notable drop in cooperative success rate from 0.550 on maze\_05 to 0.300 on maze\_mix, a cross-scenario generalisation gap of 0.250, whereas the full framework exhibits a gap of only 0.050 (0.800 to 0.750), demonstrating that the structure-aware generalisation mechanism substantially reduces policy dependence on the spatial layout of training scenarios. Standard MASAC achieves similar success rates of 0.500 and 0.450 across the two scenarios, however, both values remain substantially below the full framework, and the persistently high collision rate of 0.400 in both scenarios indicates that the policy relies on indiscriminate exploratory behaviour rather than structured navigation, yielding comparably low-level performance regardless of scenario geometry. Furthermore, MAPPO records a collision rate of 0.350 on maze\_mix, considerably higher than the full framework (0.100), indicating that without execution-level local optima intervention and safety-aware scheduling, flight safety in complex unseen environments cannot be reliably maintained.

\begin{table}[t]
\centering
\caption{Comparison results on maze\_05 and the unseen test scenario maze\_mix (20 episodes each). $\eta$: cooperative success rate; $\xi$: collision rate; Dist.: mean terminal distance.}
\label{tab:comparison}
\renewcommand{\arraystretch}{1.15}
\begin{tabular}{llccc}
\toprule
\textbf{Scenario} & \textbf{Method} & $\eta$ & $\xi$ & \textbf{Dist.\ (m)} \\
\midrule
\multirow{3}{*}{maze\_05}
 & MAPPO          & 0.550 & 0.200 &  2.910 \\
 & MASAC & 0.500 & 0.400 &  7.199 \\
 & Full framework & \textbf{0.800} & \textbf{0.050} & \textbf{2.510} \\
\midrule
\multirow{3}{*}{maze\_mix}
 & MAPPO          & 0.300 & 0.350 & 11.341 \\
 & MASAC & 0.450 & 0.400 &  8.151 \\
 & Full framework & \textbf{0.750} & \textbf{0.100} & \textbf{3.260} \\
\bottomrule
\end{tabular}
\end{table}

\subsection{Ablation Study}

\begin{figure}[htbp]
    \centering
    \includegraphics[width=\columnwidth]{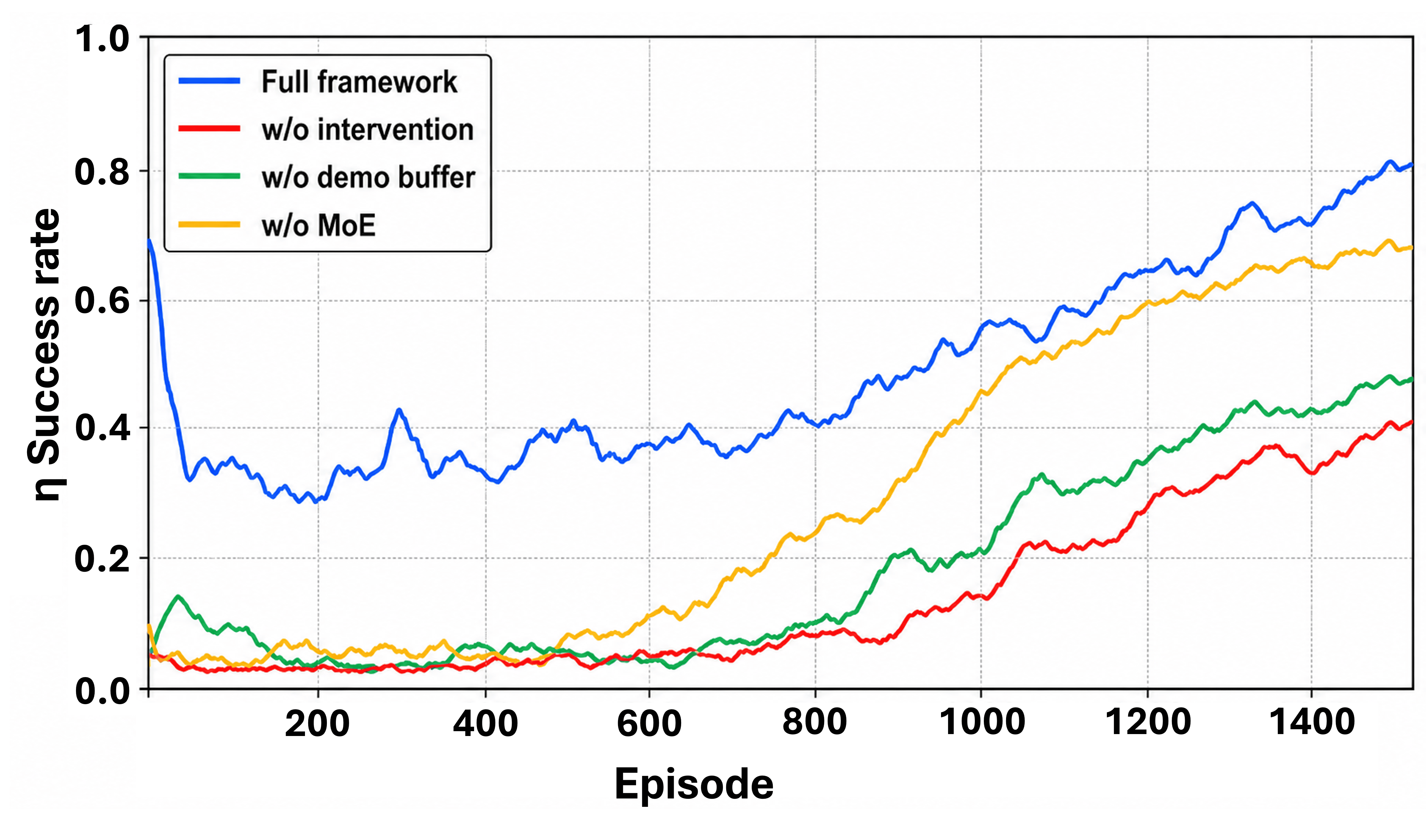}
    \caption{Cooperative success rate $\eta$ during training on maze\_05 for all ablation variants.}
    \label{fig:ablation_success}
\end{figure}

\begin{figure}[htbp]
    \centering
    \includegraphics[width=\columnwidth]{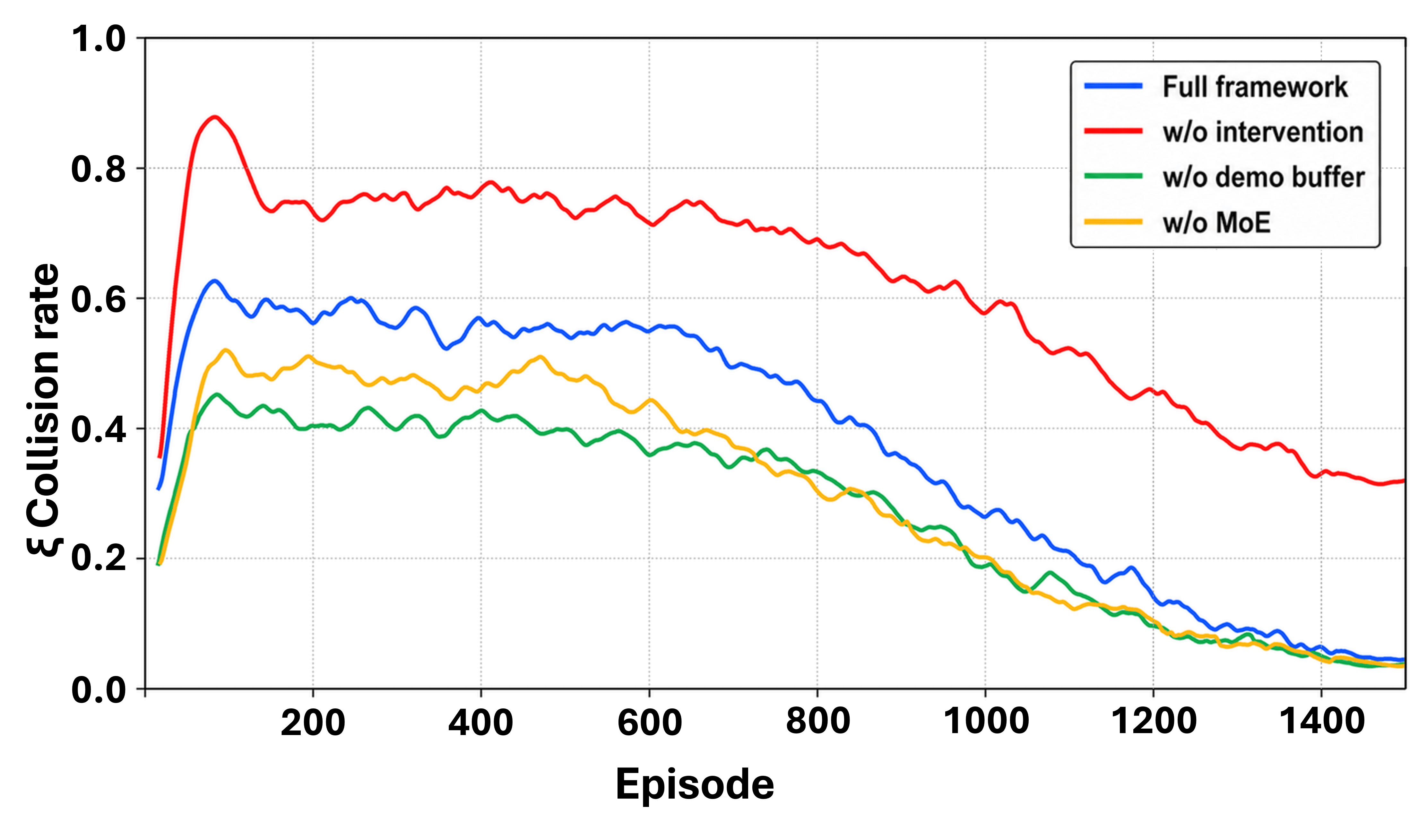}
    \caption{Collision rate $\xi$ during training on maze\_05 for all ablation variants.}
    \label{fig:ablation_collision}
\end{figure}

To evaluate the independent contribution of each proposed mechanism, ablation experiments are conducted on maze\_05 with four variants: the complete framework, w/o intervention (execution-level local optima override disabled), w/o demo (hierarchical demonstration buffer and behavioural cloning removed), and w/o MoE (MoE actor replaced by a single expert head). All variants are trained from the same random initialisation. The cooperative success rate $\eta$ and collision rate $\xi$ across training episodes are shown in Fig.~\ref{fig:ablation_success} and Fig.~\ref{fig:ablation_collision} respectively.

The full framework (blue) maintains a cooperative success rate of 0.30--0.45 from the earliest training episodes, substantially above all ablation variants, demonstrating that the graded behavioural cloning supervision from the hierarchical demonstration buffer provides effective early-stage guidance. After episode 600, $\eta$ rises steadily to approximately 0.80 by episode 1500, whilst $\xi$ concurrently falls from 0.60 to near zero, indicating that the policy converges to both high cooperative task performance and safe navigation simultaneously.

Removing the local optima intervention mechanism (w/o intervention, red) produces the most pronounced degradation. The success rate remains near zero for the first 800 episodes, whilst $\xi$ surges to approximately 0.90 within the first 100 episodes and sustains a high level of 0.70--0.80 thereafter. Without execution-level escape guidance, agents become trapped in structural dead-ends and repeatedly collide with obstacles, preventing any cooperative success signal from emerging during early exploration and delaying convergence until after episode 800, with a final $\eta$ of approximately 0.40.

Removing the hierarchical demonstration buffer (w/o demo, green) results in a success rate that remains near zero before episode 800, with occasional transient rises that subsequently collapse, reflecting the instability caused by the absence of behavioural cloning supervision. Notably, w/o demo exhibits a consistently lower collision rate than the other ablation variants across most of the training process, indicating that without cloning constraints the policy adopts conservative exploratory behaviour to avoid collisions at the cost of severely sparse cooperative success signals, substantially reducing sample efficiency relative to the full framework.

Removing the MoE actor (w/o MoE, orange) yields a final $\eta$ of approximately 0.70 on the single maze\_05 training scenario, converging later than the full framework but earlier than the w/o intervention and w/o demo variants. This suggests that a single-head policy retains meaningful learning capacity within a single training scenario, and that the contribution of the MoE architecture on maze\_05 manifests primarily as faster convergence rather than a difference in asymptotic performance. However, as demonstrated in Section~\ref{sec:generalisation}, the w/o MoE variant achieves a cooperative success rate of only 0.250 on the unseen test scenario maze\_mix, substantially below the full framework's 0.750, confirming that the structure-aware gating architecture is the critical component for cross-scenario generalisation, a contribution that single-scenario ablation alone is insufficient to reveal.

\subsection{Structure-aware Generalisation}
\label{sec:generalisation}

\begin{figure}[t]
    \centering
    \includegraphics[width=\columnwidth]{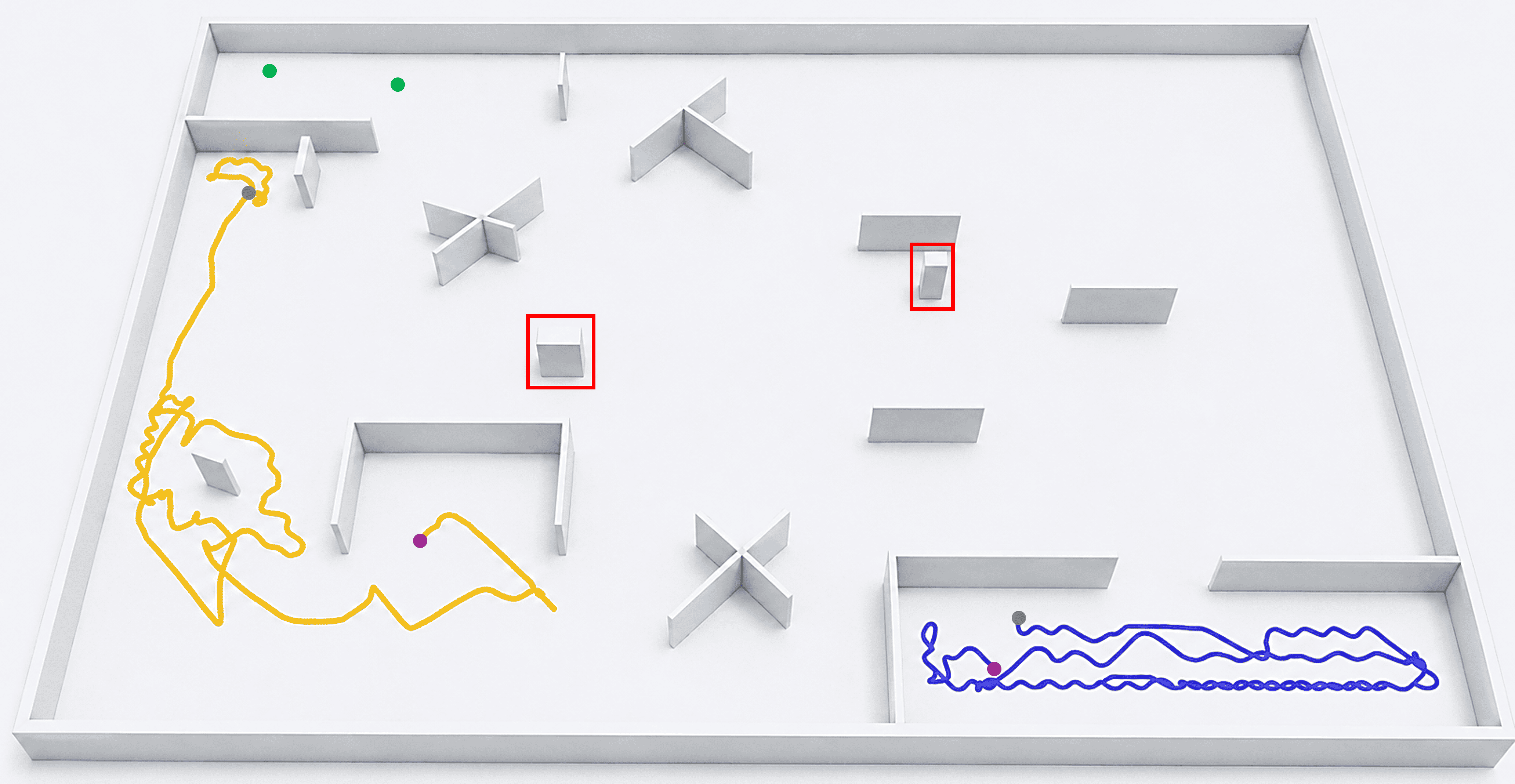}
    \caption{UAV flight trajectories of the w/o MoE variant in the unseen test scenario maze\_mix. Orange and blue curves denote UAV~1 and UAV~2 respectively; purple markers indicate start positions, green markers indicate goal positions, grey markers indicate actual terminal positions, and red bounding boxes indicate moving obstacles.}
    \label{fig:traj_wo_moe}
\end{figure}

Table~\ref{tab:generalisation} compares the proposed framework against the w/o MoE variant on the unseen test scenario maze\_mix. The full framework achieves a cooperative success rate of 0.750 under dynamic obstacle conditions throughout all episodes, with a mean terminal distance of 3.260\,m. The w/o MoE variant replaces the MoE actor with a single expert head whilst retaining $\omega$ as part of the observation input; however, its cooperative success rate drops to 0.250 and mean terminal distance rises to 17.608\,m. A representative failure episode is illustrated in Fig.~\ref{fig:traj_wo_moe}: UAV~1 becomes trapped in the local optima structure located in the upper-left region ahead of its goal, whilst UAV~2 remains confined to the lower-right corridor and fails to identify a viable navigation path. Although UAV~1 reaches its goal individually in 14 out of 20 episodes (partial success rate 0.700) and the team achieves full cooperative success in 5 episodes ($\eta$=0.250), UAV~2 records a mean terminal distance of 31.49\,m across all episodes, demonstrating that a single-head policy cannot simultaneously provide effective navigational responses to two UAVs traversing structurally distinct local environments. These results indicate that incorporating $\omega$ solely as an observation feature is insufficient for cross-scenario transfer; the structure-specialised routing provided by the MoE gating architecture is the critical factor underlying generalisation capability.

Analysis of the gate weight distributions of the full framework further validates the working mechanism. The mean activation weights of the four expert heads for UAV~1 are: dead-end expert 0.000, wall expert 0.026, narrow expert 0.000, and open expert 0.974, indicating that the open expert dominates its navigation path, reflecting a route predominantly traversing open areas, whilst the marginal activation of the wall expert suggests the presence of occasional single-sided wall structures along the way. UAV~2 exhibits a markedly different activation distribution: dead-end expert 0.036, wall expert 0.065, narrow expert 0.346, and open expert 0.553, with all four expert heads showing non-negligible activation, amongst which the narrow expert and open expert jointly dominate, indicating that its path encompasses both open areas and narrow passages, requiring the gating network to dynamically switch expert responses across distinct local structures. This automatic divergence in expert activation between the two UAVs is not manually prescribed but emerges from the gating network computing $\omega$ independently from each UAV's real-time LiDAR readings, demonstrating that the structure-aware gating network performs fine-grained adaptive responses to the local structural differences experienced by individual agents within the same unseen scenario, rather than memorising the global spatial layout of training environments.

\begin{table}[t]
\centering
\caption{Cross-scenario transfer results on the unseen test scenario maze\_mix (20 episodes). $\eta$: cooperative success rate; $\xi$: collision rate; Dist.: mean terminal distance.}
\label{tab:generalisation}
\renewcommand{\arraystretch}{1.15}
\begin{tabular}{lccc}
\toprule
\textbf{Variant} & $\eta$ & $\xi$ & \textbf{Dist.\ (m)} \\
\midrule
Full framework & 0.750 & 0.100 & 3.260 \\
w/o MoE        & 0.250 & 0.050 & 17.608 \\
\bottomrule
\end{tabular}
\end{table}

\section{Conclusion}

This paper proposes a multi-agent deep reinforcement learning framework for cooperative UAV navigation in complex constrained environments, systematically addressing local optima traps, sparse cooperative reward signals, inter-agent learning imbalance, and insufficient cross-scenario generalisation through four coordinated mechanisms.

Simulation experiments validate the effectiveness of the proposed framework and each of its constituent mechanisms. The complete framework passes the curriculum advancement thresholds across all static maze training scenarios and maintains stable cooperative navigation performance in dynamic obstacle environments. On an unseen mixed test scenario not encountered during training, the framework achieves effective zero-shot cross-scenario transfer, validating the structure-aware generalisation mechanism in practice. Comparison experiments against existing multi-agent reinforcement learning baselines demonstrate that the full framework achieves superior cooperative success rate and collision safety on both the training scenario and the unseen test scenario, with a substantially smaller cross-scenario generalisation gap than the baseline methods, confirming the overall advantage of the proposed approach. Ablation experiments further demonstrate that the local optima intervention mechanism is decisive for training efficiency during early exploration, that the hierarchical demonstration buffer and graded behavioural cloning supervision substantially improve sample utilisation efficiency under sparse cooperative success signals, and that the MoE gating architecture is the critical component underlying cross-scenario generalisation capability, a contribution that single-scenario ablation alone is insufficient to reveal and that the generalisation experiments provide more direct evidence for.

The current study is validated under a two-UAV configuration, and the gap between simulation and real-world deployment remains to be bridged. Future work will proceed in three directions: extending the framework to larger multi-agent teams; validating deployment on real UAV platforms; and exploring the learning of shared local geometric features across a broader range of constrained environment types, with the aim of further improving the framework's adaptability to unseen environments.

\section*{Acknowledgment}
The authors used Claude for English language proofreading during the preparation of this manuscript.

\bibliographystyle{IEEEtran}
\bibliography{refs}

\end{document}